\pdfoutput=1

\documentclass[11pt]{article}

\usepackage[preprint]{acl}

\usepackage{times}
\usepackage{latexsym}

\usepackage[T1]{fontenc}
\usepackage[utf8]{inputenc}

\usepackage{microtype}
\usepackage{inconsolata}

\usepackage{graphicx}
\usepackage{multirow} 
\usepackage{caption}
\usepackage{booktabs}  
\usepackage{makecell}
\usepackage{array}
\usepackage{stfloats}
\usepackage{float}
\usepackage{algorithm}
\usepackage{algorithmic} 
\usepackage{amsmath,amssymb} 
\usepackage{adjustbox}
\usepackage[most]{tcolorbox}
\usepackage{booktabs}
\usepackage{tablefootnote}
\usepackage{tcolorbox}

\title{\textit{“Well, Keep Thinking”}: Enhancing LLM Reasoning\\with Adaptive Injection Decoding}

\author{%
Hyunbin Jin\thanks{\ \ These authors contributed equally to this work.} \quad
Je Won Yeom\footnotemark[1] \quad
Seunghyun Bae\footnotemark[1] \quad
Taesup Kim\thanks{\ \ Corresponding author.} \\
Graduate School of Data Science, Seoul National University\\
\texttt{\{hyunbin.jin, jewon0908, sh.bae, taesup.kim\}@snu.ac.kr}
}

\begin{document}
\maketitle

\begin{abstract}
    Large language models (LLMs) exhibit strong reasoning abilities, often attributed to few-shot or zero-shot chain-of-thought (CoT) prompting.
While effective, these methods require labor-intensive prompt engineering, raising the question of whether reasoning can be induced without reliance on explicit prompts.
In this work, we unlock the reasoning capabilities of LLMs without explicit prompting.
Inspired by zero-shot CoT and CoT-decoding, we propose a novel decoding strategy that systematically nudges LLMs to continue reasoning, thereby preventing immature reasoning processes. 
Specifically, we monitor the model’s generation and inject a designated phrase whenever it is likely to conclude its response prematurely, before completing the reasoning process. 
Our experimental evaluations on diverse reasoning benchmarks demonstrate that our proposed strategy substantially improves LLM reasoning capabilities, highlighting the potential of decoding-based interventions as an alternative to traditional prompting techniques.
\end{abstract}

\section{Introduction}


Large Language Models (LLMs) have revolutionized natural language processing (NLP), demonstrating remarkable performance across a wide range of tasks, including question-answering, code generation, and machine translation.
Scaling up model size has led to improvements in fluency, generalization, and sample efficiency~\cite{wei2022emergent}.
However, despite these advancements, LLMs continue to struggle with complex reasoning tasks such as arithmetic, commonsense reasoning, and multi-step logical inference.
To address these challenges, researchers have primarily relied on prompt engineering techniques, such as few-shot Chain-of-Thought (CoT) prompting and zero-shot CoT~\citep{brown2020languagemodelsfewshot, besta2023graph, madaan2023selfrefine, wei2022chainofthought, yao2023tree}.
These approaches guide LLMs by providing structured reasoning examples or explicit step-by-step instructions, significantly improving performance on various reasoning benchmarks.
However, they exhibit key limitations in that few-shot prompting requires human-crafted exemplars, while zero-shot prompting is highly sensitive to prompt phrasing.
More recently, some studies have explored CoT reasoning without prompts by modifying the decoding strategy~\citep{wang2023cotwithout, wang2023selfconsistency}. 
Additionally, a test-time scaling method has been introduced, which intervenes during inference to control test-time compute~\citep{muennighoff2025s1}. 

\begin{figure}[t!]
\centering
  \includegraphics[width=\linewidth]{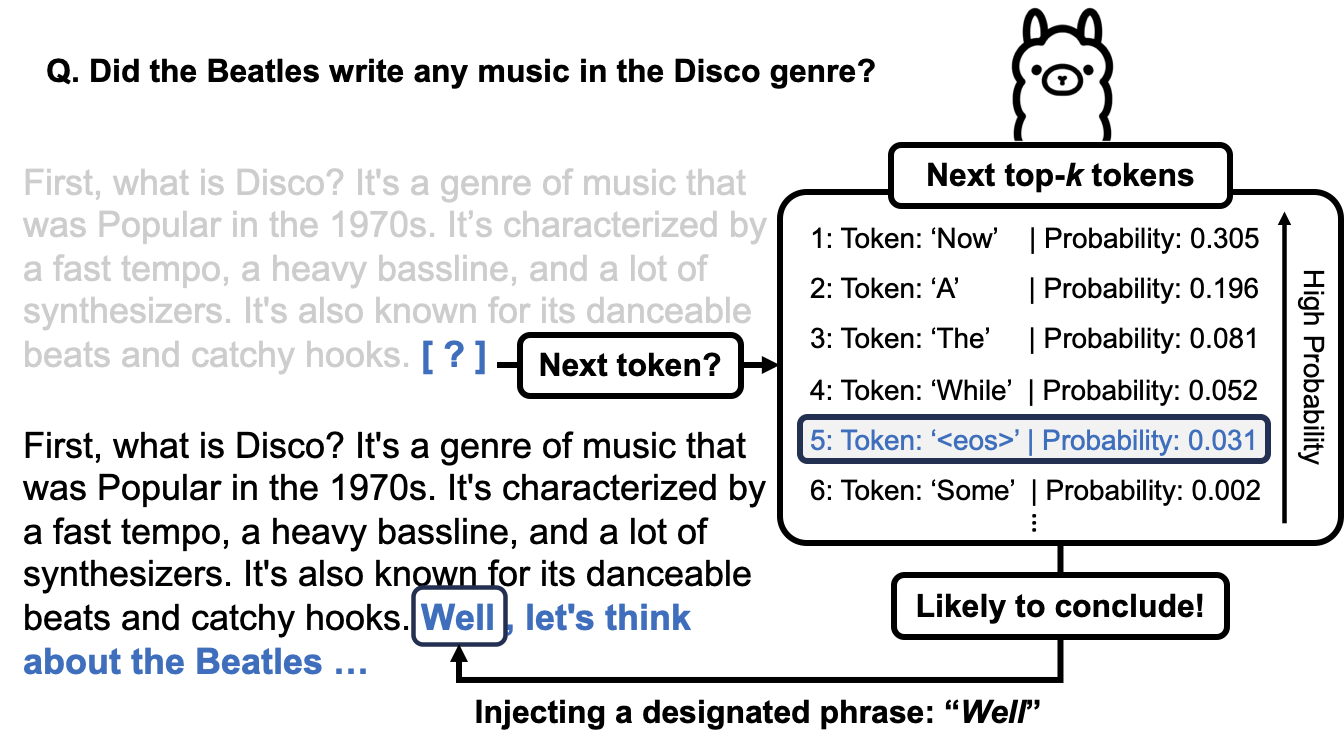}
  \caption{Overview of our proposed method, \emph{Adaptive Injection Decoding}, with injecting a designated phrase ``\texttt{Well}'' whenever the model is likely to conclude its response immaturely with a \texttt{<eos>} token.}
  \label{fig:example}
\end{figure}

Despite various efforts to enhance LLM reasoning, existing approaches have not directly addressed the fundamental causes of their reasoning failures. Through our analysis, we categorize these failures into three primary types, highlighting the need for a more targeted intervention.
First, models terminate generation without producing a response to given questions, returning only the <eos> token, a phenomenon we refer to as \textit{silence}. Second, models often produce responses that lack actual logical reasoning, responding with uninformative outputs or irrelevant patterns, such as numeric sequences or question repetition, a failure type we call \textit{no reasoning}. Finally, in some cases, models initiate reasoning but halt prematurely, leading to incomplete or logically inconsistent answers, which we refer to as \textit{incomplete reasoning}.
These failure types suggest that LLMs do not inherently lack reasoning ability, rather  their reasoning processes are disrupted or prematurely terminated during decoding.

To address these reasoning failures, we posed a fundamental question: {\textit{“How do humans think effectively when solving complex problems? What strategies help people avoid reasoning failures, and how do they enrich their thought process to reach accurate conclusions?”}}.
When solving complex problems, humans rarely produce an immediate answer.
Instead, they pause, reflect, and often restart their thought process before reaching a final conclusion.
When encountering difficulties, people frequently engage in self-prompting mechanisms, such as asking themselves guiding questions, articulating partial thoughts, or seeking alternative perspectives to expand their reasoning. 
These cognitive strategies ensure that reasoning continues rather than stopping prematurely, allowing for deeper and more thorough problem-solving.

Inspired by this cognitive process, we propose a novel test-time intervention technique based on an adaptive decoding strategy.
Rather than fully relying on a greedy decoding strategy, our method dynamically intervenes with a designated injection phrase during inference when the model's reasoning process is on the verge of prematurely terminating or drifting away from logical reasoning.

\begin{figure*}[t]
\centering
  \includegraphics[width=\textwidth]{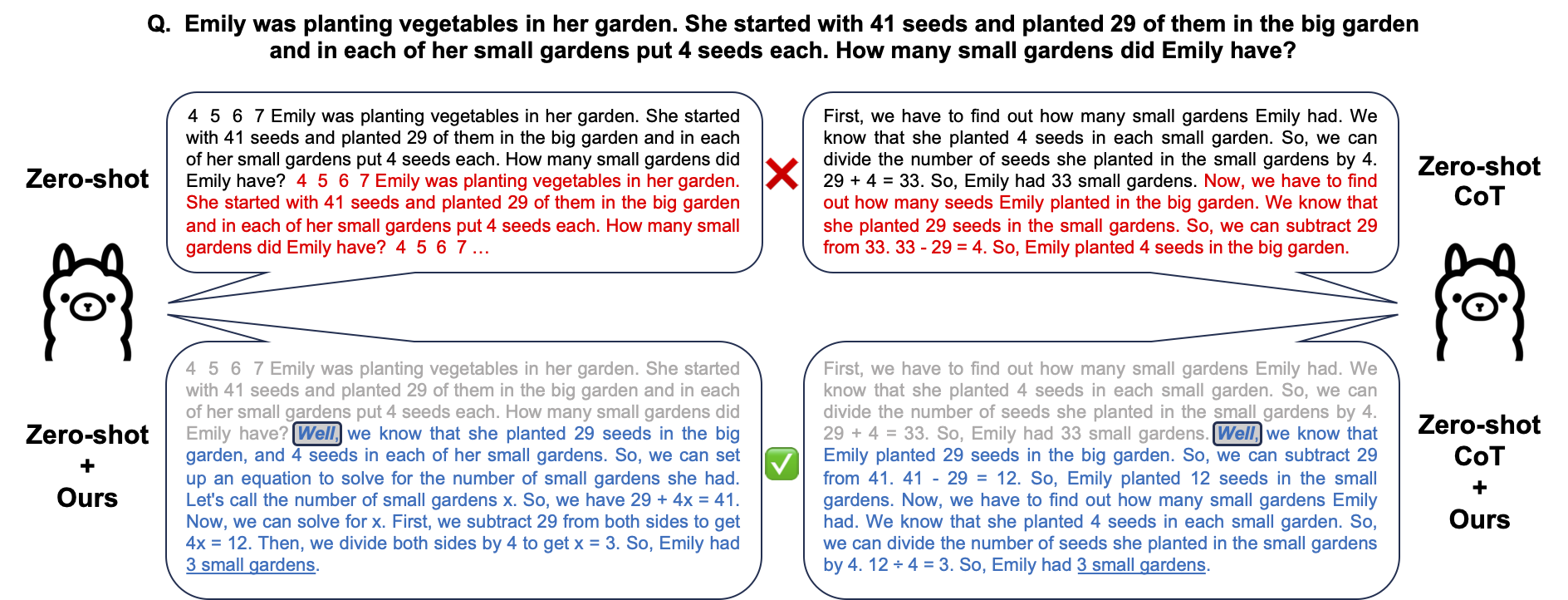}
  \caption{Example question and generated answers of Llama-3.1-8B with (a) Standard Zero-shot, (b)
Zero-shot + Ours, (c) Zero-shot-CoT, and (d) Zero-shot-CoT + Ours settings.}
  \label{fig:example}
\end{figure*}

Similar to how humans engage in self-reflection to sustain their reasoning process, our adaptive injection decoding strategy serves as an \textit{implicit signal} that prevents premature termination and encourages adequate reasoning. 
Our contributions are as follows:
\begin{itemize}
    \item We identify the \textit{immature reasoning} as a major limitation in LLMs and systematically classify common reasoning failures.
    \item We propose a novel test-time intervention that extends reasoning dynamically with an \textit{adaptive injection decoding} strategy, ensuring logical continuity.
    \item Our approach can significantly improve reasoning performance across reasoning benchmarks and is computationally efficient, making it suitable for practical real-world applications.
\end{itemize}

\section{Observations: Immature Reasoning}
\label{sec:observations}
We conducted an experiment using LLaMA 3.1-8B on the GSM8K dataset to assess the reasoning performance of LLMs and identify the underlying factors contributing to their suboptimal performance. 
Our analysis highlights several challenges affecting the model’s reasoning abilities while also revealing insights into its latent potential. 
The following sections summarize key observations, with specific examples provided in Appendix \ref{sec:premature-termination}.

\begin{figure}[t!]
\centering
  \includegraphics[width=\columnwidth]{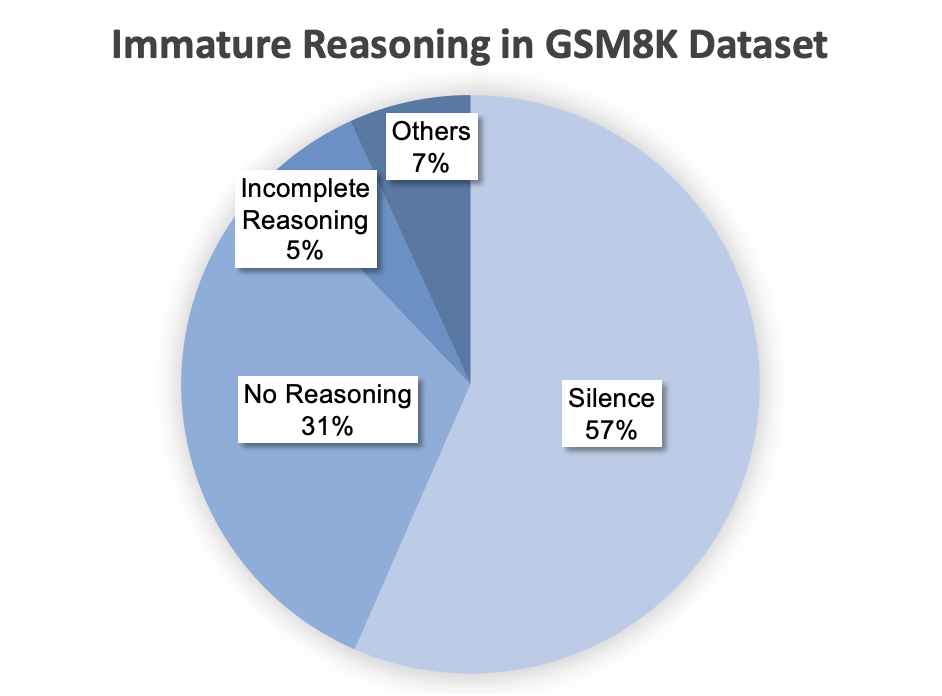} 
  \caption{Distribution of \textit{immature reasoning} categories for incorrect answers in GSM8K using LLaMA 3.1-8B}
  \label{fig:observation-category}
\end{figure}

We identify \textit{Immature Reasoning} as a major factor behind the poor reasoning performance of LLMs. 
Immature reasoning refers to instances where the model prematurely halts or drifting away from logical reasoning process, leading to incomplete, irrelevant, or absent responses. 
Our analysis of the model's responses revealed three distinct types of immature reasoning.
\paragraph{(Observation 1) \textit{Silence}} 
The model fails to generate any response despite receiving an input question.
This typically occurs when the model gives up on reasoning and refrains from attempting an answer.
In our evaluation of LLaMA 3.1-8B on GSM8k, 58.05\% of incorrect responses fell into this category.
Refer to Appendix \ref{sec:silence} for examples of this category.
\paragraph{(Observation 2) \textit{No Reasoning}}
The model generates a response but does not engage in actual reasoning.
Instead, it follows learned response patterns, repeats the question, or outputs irrelevant content (e.g., numeric sequences or HTML) without attempting to solve the problem logically. 
This issue accounted for 31.7\% of incorrect responses.
Refer to Appendix \ref{sec:no-reasoning} for examples of this category.
\paragraph{(Observation 3) \textit{Incomplete Reasoning}} 
The model initiates reasoning but stops short of completing the necessary steps to solve the problem.
It may prematurely terminate its reasoning process or deviate from the original problem-solving task, focusing on irrelevant examples or formats.
This type of failure represented 5.45\% of incorrect responses.
Refer to Appendix \ref{sec:incomplete-reasoning} for examples of this category.

Despite these challenges, we observed that a substantial portion of responses that exhibited appropriate reasoning successfully arrived at correct solutions.
In our evaluation of LLaMA 3.1-8B on GSM8k, 32.22\% of responses that engaged in proper CoT reasoning yielded correct answers.
This suggests that LLMs are not inherently deficient in problem-solving capabilities.
Instead, immature reasoning disrupts the reasoning process, preventing models from fully realizing their potential.



\begin{algorithm}[t!]
\centering
\small
\caption{\textbf{Adaptive Injection Decoding}}
\label{alg:topk}
\begin{algorithmic}[1]

\REQUIRE model $f(\cdot)$, prompt $x$, top-$k$ $k$, injection prompt $p$, using zero-shot-CoT \texttt{zs\_cot}, maximum length \texttt{max\_length}

\IF{ \texttt{zs\_cot} }
  \STATE $x \gets x + \text{"}\backslash n\text{Let's think step by step."}$ 
\ENDIF
\STATE $r \gets [\,]$; 
\STATE $\texttt{cont} \gets \text{False}$

\FOR{$i = 1$ to \texttt{max\_length}}
  \STATE $\texttt{p} \gets f(x)$ 
  \STATE \texttt{top\_k} $\gets$ \text{TopK}$(\texttt{p}, k)$
  \IF{\textbf{not }\texttt{cont} \textbf{ and } \texttt{<eos>} $\in$ \texttt{top\_k}}
    \STATE $x \gets x + p$
    \STATE $r \gets r + p$
    \STATE $\texttt{cont} \gets \text{True}$
    \STATE \textbf{continue}
  \ENDIF

  \STATE \texttt{next\_token} $\gets \arg\max(\texttt{p})$
  \IF{\texttt{next\_token}  is \texttt{<eos>}}
    \STATE \textbf{break}
  \ENDIF

  \STATE $x \gets x + $ \texttt{next\_token}
  \STATE $r \gets r + $ \texttt{next\_token}
\ENDFOR

\RETURN $r$

\end{algorithmic}
\label{alg:algorithm}
\end{algorithm}

\section{Adaptive Injection Decoding}
To mitigate \textit{immature reasoning}, we introduce a novel decoding strategy. 
Our approach, \textit{Adaptive Injection Decoding (AID)}, dynamically intervenes in such cases by injecting a predefined \textit{injection phrase} (e.g., \texttt{``Well''}) as the next token whenever the probability of generating the \texttt{<eos>} token meets a specified criterion.
This method is designed to prevent the model from prematurely concluding its reasoning process or deviating logical reasoning.

Given an input query $x$ as an input prompt, the model $f$ generates a sequence of response tokens $r_1, r_2, \dots, r_t$ through greedy decoding. 
At each decoding step $t$, if the \texttt{<eos>} token ranks among the top-$k$ next token predictions, it is replaced with injection phrase $p$. 
The updated response $r_1, r_2, \dots, r_t, p$ is then fed back into the model, ensuring continuity of reasoning. 
The algorithmic details of our method are provided in Algorithm~\ref{alg:algorithm}.

To properly prevent premature termination and encourage reasoning, we rigorously evaluated various injection phrases, including single words, conjunctions, full sentences, and machine language expressions. 
Among these, \texttt{``Well''} consistently outperformed other alternatives, as shown in Table \ref{tab:main-token-ablation}, and was selected for all subsequent experiments.
Further details on the experimental analysis are discussed later in Section~\ref{ablation}.

Additionally, our experiments revealed distinct model-specific tendencies in handling the \texttt{<eos>} token, particularly in how early and strongly it is considered during generation. LLaMA-3.1-8B tends to terminate responses earlier, whereas Mistral-7B-v0.3 and Gemma-7B generate longer outputs. However, despite their delayed termination, Mistral-7B-v0.3 and Gemma-7B consistently ranked \texttt{<eos>} among the top candidates from the early stages of generation. This suggests that while they had a higher potential for early termination, alternative tokens were ultimately selected. 
Building on this observation, we explored replacing injection phrases when \texttt{<eos>} appeared in the top-$k$ candidates and further investigated optimal top-$k$ values for each model in subsequent experiments (see Section~\ref{ablation} for details).

\section{Experiments}
\subsection{Experimental Setup}
\paragraph{Models}
We mainly investigate LLMs, including LLaMA-3.1-8B, Mistral-7B-v0.3, and Gemma-7B~\cite{touvron2023llama, siino-2024-mistral, mesnard2024gemma}. 

\paragraph{Tasks and Datasets}
We evaluate our proposed method across a variety of reasoning benchmarks, including tasks such as arithmetic reasoning, commonsense reasoning and logical reasoning.
(a) \textit{Arithmetic Reasoning:} {MultiArith} \citep{roy2015solving} includes elementary school arithmetic questions. {GSM8K} \citep{cobbe2021training} consists of grade-school-level math word problems requiring multi-step reasoning. 
(b) \textit{Commonsense Reasoning:} {StrategyQA} \citep{geva2021didaristotle} includes yes/no questions that require multi-step, strategy-based reasoning. {BBH-Disambiguation QA} \citep{suzgun2023challenging} consists of questions on pronoun resolution and ambiguity detection.
(c) \textit{Logical Reasoning:} {BBH-Logical Deduction} (Five Objects) \citep{suzgun2023challenging} includes questions involving logical deduction.

\paragraph{Baselines}
We compare our method with the following baselines. (a) \textit{Zero-shot (i.e., Greedy Decoding)}: The model generates responses without any intervention, guidance, or modification.
(b) \textit{Zero-shot-CoT (Prompting)}: An initial prompt is added to the input to facilitate CoT reasoning. Specifically, we use the prompt \texttt{``Let’s think step by step''} to guide progressive problem-solving.


\paragraph{LLM-based Evaluation}
To assess the accuracy of the generated responses, we utilize \texttt{o1-mini} via OpenAI’s chat completion API (version 2024-09-12) \citep{openai2024o1mini}. The evaluation prompt template, including grading criteria and methodology, is provided in Appendix~\ref{sec:evaluation_prompt}. We also compare \texttt{o1-mini} and \texttt{gpt-4o-mini} with human evaluations across multiple datasets using LLaMA-3.1-8B (Table~\ref{tab:alignment}), where \texttt{o1-mini} shows markedly closer alignment to human judgment. Further details are in Appendix~\ref{sec:evaluation_prompt}.




\begin{table*}[t!]
\centering
\resizebox{\textwidth}{!}{
\begin{tabular}{c|l|cc|cc|c|c} \toprule
\multicolumn{1}{c}{\multirow{2}{*}{Model}} & \multicolumn{1}{c}{\multirow{2}{*}{Method}} & \multicolumn{2}{c}{Arithmetic} & \multicolumn{2}{c}{Commonsense} & \multicolumn{1}{c}{Logical} & \multirow{2}{*}{Avg} \\ \cmidrule(l){3-6}\cmidrule(lr){7-7}
\multicolumn{1}{c}{} & \multicolumn{1}{c}{} & MultiArith & \multicolumn{1}{c}{GSM8K} & StrategyQA & \multicolumn{1}{c}{DisambiguationQA} & \multicolumn{1}{c}{Logical Deduction} &  \\ \midrule\midrule
\multirow{4}{*}{Llama-3.1-8B} & Zero-shot & 15.56\% & 6.97\% & 26.35\% & 36.00\% & 28.40\% & 22.66\% \\
 & Zero-shot + Ours & \textbf{50.56\%} & \textbf{34.57\%} & \textbf{30.13}\% & \textbf{37.20\%} & \textbf{32.00\%} & \textbf{36.90\%} \\
 & Zero-shot-CoT & 77.22\% & \textbf{48.90\%} & 24.31\% & 32.00\% & 16.80\% & 39.85\% \\
 & Zero-shot-CoT + Ours & \textbf{78.33\%} & 34.34\% & \textbf{45.27}\% & \textbf{34.00\%} & \textbf{24.80\%} & \textbf{43.35\%} \\ \midrule
\multirow{4}{*}{Gemma-7b} & Zero-shot & 10.00\% & \textbf{48.98\%} & 31.88\% & 10.80\% & 24.80\% & 25.29\% \\
 & Zero-shot + Ours & \textbf{15.00\%} & 45.26\% & \textbf{34.79\%} & \textbf{16.80\%} & \textbf{27.20\%} & \textbf{27.81\%} \\
 & Zero-shot-CoT & 46.11\% & 38.59\% & 19.36\% & 34.00\% & 18.80\% & 31.37\% \\
 & Zero-shot-CoT + Ours & \textbf{73.33\%} & \textbf{44.81\%} & \textbf{23.44\%} & \textbf{36.00\%} & \textbf{24.00\%} & \textbf{40.32\%} \\ \midrule
\multirow{4}{*}{Mistral-7B-v0.3} & Zero-shot & 16.11\% & 7.73\% & 23.58\% & 34.40\% & 14.80\% & 19.32\% \\
 & Zero-shot + Ours & \textbf{27.78\%} & \textbf{15.31\%} & \textbf{25.62\%} & \textbf{37.20\%} & \textbf{18.00\%} & \textbf{24.78\%} \\
 & Zero-shot-CoT & 29.44\% & \textbf{17.21\%} & 16.30\% & 13.20\% & \textbf{28.80\%} & 20.99\% \\
 & Zero-shot-CoT + Ours & \textbf{37.22\%} & 15.39\% & \textbf{36.83\%} & \textbf{32.00\%} & 20.40\% & \textbf{28.37\%} \\ \bottomrule
\end{tabular}
}
\caption{Accuracy (\%) of LLaMA-3.1-8B, Mistral-7B-v0.3, and Gemma-7B models across a wide range of reasoning tasks. We compared four settings: Zero-shot, Zero-shot + Ours, Zero-shot-CoT, and Zero-shot-CoT + Ours.}
\label{tab:model-ablation}
\end{table*}
\subsection{Results}
Table \ref{tab:model-ablation} presents the performance of our method applied to both zero-shot and zero-shot-CoT baselines.
Our approach consistently improves performance across all three models, enhancing results on a wide range of tasks, including arithmetic, commonsense, and logical reasoning. 
These improvements are reflected in the \textbf{Avg} column of Table~\ref{tab:model-ablation}.

\paragraph{Arithmetic Reasoning}
In MultiArith, each model achieves at least a 150\% increase in accuracy simply by injecting our proposed phrase \texttt{“Well”} during inference, even without using zero-shot-CoT prompting. 
For instance, in LLaMA-3.1-8B, our decoding strategy increases accuracy from 15.56\% to 50.56\%, with an additional prompt (e.g., “Let’s think step by step”) further improving performance to 78.33\%. A similar trend is observed in Gemma-7B, where accuracy rises from 10.00\% to 15.00\%, and then to 73.33\% with the inclusion of the additional prompt. Similarly, Mistral-7B-v0.3 demonstrates an improvement from 16.11\% to 27.78\%, ultimately achieving an accuracy of 37.22\%.


\paragraph{Commonsense and Logical Reasoning}
Along with arithmetic reasoning, we also observe significant improvements in commonsense and logical reasoning tasks.
For instance, Mistral-7B-v0.3 achieves a +20.53\% performance gain (16.30\% → 36.83\%) in StrategyQA and a +18.80\% increase (13.20\% → 32.00\%) in DisambiguationQA under zero-shot-CoT with our method. Similarly, LLaMA-3.1-8B improves from 16.80\% to 24.80\% in Logical Deduction under zero-shot-CoT, while Gemma-7B increases from 18.80\% to 24.00\% under the same conditions. These results highlight that even simple intervention during decoding can yield substantial performance gains across a wide range of reasoning tasks.

\paragraph{Enhancing Prompt-based Reasoning}
Our method integrates seamlessly with prompting techniques, consistently improving performance compared to using prompting alone.
As shown in Table~\ref{tab:model-ablation}, the average performance of each model across datasets demonstrates that combining our decoding strategy with an initial prompt results in a significant enhancement over prompting alone.
Notably, this improvement is nearly twice as large as that achieved in the zero-shot setting.
For example, in MultiArith, where zero-shot prompting already provides some improvement, the application of our method results in additional performance gains.

\paragraph{Recovery from Ineffective Prompting}
In certain cases, utilizing zero-shot-CoT prompts can reduce accuracy compared to not using any prompts, as demonstrated in Table~\ref{tab:model-ablation} (e.g., StrategyQA, DisambiguityQA, Logical Deduction). This detrimental effect of zero-shot-CoT prompting has been previously reported.
\citet{chakraborty2023zero}
highlight that language model performance is highly sensitive to prompt phrasing, and manually designed prompts often lead to suboptimal results in zero-shot settings.
However, our method effectively mitigates these adverse effects, either restoring or even surpassing performance observed prior to the application of prompts. 
For instance, in StrategyQA, applying a zero-shot-CoT prompt reduces Mistral-7B-v0.3’s accuracy from 23.58\% to 16.30\%. 
In contrast, injecting our method boosts accuracy to accuracy to 36.83\%, effectively counteracting the negative effects of prompting and surpassing the model’s previous best performance on this dataset.
A similar trend is observed in Logical Deduction, further validating the effectiveness of our approach.

\begin{table}[t]
\centering
\resizebox{\columnwidth}{!}{
\begin{tabular}{c|c|c} \toprule
\multicolumn{1}{c}{Category} & \multicolumn{1}{c}{Injection Phrase} & Accuracy \\ \midrule\midrule
\multirow{4}{*}{Single Word} 
& Step & 44.44\% \\
 & Let & 38.33\% \\
 & \textbf{Well} & \textbf{50.56\%} \\
 & Wait~\cite{muennighoff2025s1}& 21.11\% \\
 \midrule
\multirow{3}{*}{Conjunction} 
& And & 16.11\% \\
 & But & 17.78\% \\  
 & Or & 12.78\% \\   \midrule
\multirow{3}{*}{Conjunction Pool} 
& Addition Pool & 26.11\% \\
 & Contrast Pool & 20.56\% \\
 & Mix Pool & 23.33\% \\ \midrule
\multirow{2}{*}{Phrase} 
& I mean, & 27.22\% \\
 & You know, & 22.22\% \\ \midrule
\multirow{2}{*}{Sentence} 
& I might be wrong. & 33.33\% \\
 & Keep reasoning. & 12.22\% \\ \midrule
 \multirow{3}{*}{Machine Language} 
& \string<start of text> & 27.78\% \\
 & \textbackslash{}t & 19.44\% \\  
 & \# & 26.67\% \\  
 \bottomrule

\end{tabular}
} 
\caption{
{Injection Phrase Analysis.} Accuracy of different injection phrases in the MultiArith dataset, evaluated on the LLaMA-3.1-8B model.
}
\label{tab:main-token-ablation}

\end{table}

\subsection{Ablation Study and Analysis}
\label{ablation}
\paragraph{Injection Phrase Analysis} Building on our findings in Section~\ref{sec:observations}, we now examine how different candidate phrases can help prevent abrupt termination and encourage more complete reasoning. The goal is to nudge the model toward continuing its chain of thought, rather than concluding too soon.

To compare how different injection words or phrases influence performance, we conducted a comprehensive ablation study, as shown in Table \ref{tab:main-token-ablation}.
Among all tested options, “\texttt{Well}” proved to be the most consistently effective and was selected for our experiments.

We interpret “\texttt{Well}” stands out because it provides a neutral yet effective nudge, encouraging the model to continue reasoning without introducing restrictive biases. Unlike phrases like “\texttt{I might be wrong.}” or “\texttt{Keep reasoning,}” which explicitly direct the model’s confidence, “\texttt{Well}” does not steer the reasoning in a particular direction. It also avoids forcing a specific reasoning trajectory, unlike conjunction words, and does not impose a rigid structured reasoning path, unlike step-indicating words. By sidestepping these constraints, “\texttt{Well}” serves as a subtle yet impactful cue, fostering deeper reasoning while preserving flexibility across various problem types.

Our finding aligns with prior work showing that large language models are highly sensitive to short prompts or cues, including those that subtly convey the need for additional reasoning steps~\citep{kim-etal-2024-will}. Further details and examples of our injection-phrase experiments are provided in Appendix~\ref{sec:phrase_ablation}.

\begin{figure}[t] \centering \centerline{\includegraphics[width=\columnwidth]{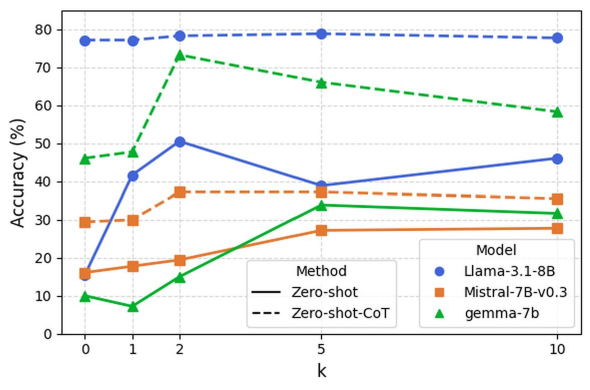}} \vspace{-1em} \caption{{The Effect of $k$ in AID.} We report zero-shot and zero-shot-CoT accuracy on the MultiArith dataset while adjusting $k$ (top-$k$) in AID . Here,  $k$ = 0  represents the case where our method is not applied.} \label{fig:top-k-ablation}
\end{figure}

\paragraph{The Effect of $k$ in AID}
During our experiments, we found that the prevalence of the \texttt{<eos>} token among top-ranked outputs varied substantially across models.
To explore this behavior, we conducted an ablation study on the MultiArith dataset, adjusting the value of $k$ for each model and measuring accuracy under zero-shot and zero-shot-CoT conditions. 
Here, $k=0$ indicates that our method is not applied. 
As shown in Figure~\ref{fig:top-k-ablation}, LLaMA-3.1-8B achieved its best performance at $k=2$.
Meanwhile, Gemma-7B and Mistral-7B-v0.3 exhibited varying preferences for either $k=2$ or $k=5$, depending on whether zero-shot-CoT was enabled.
After examining these trends, we settled on $k=2$ for Gemma-7B  and $k=5$ for Mistral-7B-v0.3, keeping these settings consistent in all subsequent experiments.

\begin{table*}[t!]
\centering
\resizebox{\textwidth}{!}{
\begin{tabular}{l|cccc|cccc|cl} \toprule
\multicolumn{1}{c}{\multirow{2}{*}{Method}} & \multicolumn{4}{c}{Arithmetic} & \multicolumn{4}{c}{Commonsense} & \multicolumn{2}{c}{Logical} \\ 
\cmidrule(l){2-11}
\multicolumn{1}{c}{} & MultiArith & GSM8K & MATH & \multicolumn{1}{c}{AQUA} & CSQA & StrategyQA & BBH-Date & \multicolumn{1}{c}{BBH-Disamb} & BBH-Logic & BBH-Web \\ \midrule\midrule
Zero-shot & 15.56\% & 6.97\% & 4.00\% & 15.35\% & \textbf{65.27\%} & 26.35\% & \textbf{22.00\%} & 36.00\% & 28.40\% & 31.20\% \\
Zero-shot + Ours & \textbf{50.56\%} & \textbf{34.57\%} & \textbf{12.80\%} & \textbf{24.41\%} & 63.88\% & \textbf{30.13\%} & 20.80\% & \textbf{37.20\%} & \textbf{32.00\%} & \textbf{34.80\%} \\ \midrule
Zero-shot-CoT & 77.22\% & \textbf{48.90\%} & \textbf{12.00\%} & \textbf{29.13\%} & 49.55\% & 24.31\% & 37.20\% & 32.00\% & 16.80\% & 14.80\% \\
Zero-shot-CoT + Ours & \textbf{78.33\%} & 34.34\% & 11.60\% & 28.74\% & \textbf{51.84\%} & \textbf{45.27\%} & \textbf{44.80\%} & \textbf{34.00\%} & \textbf{24.80\%} & \textbf{28.00\%} \\ 
\bottomrule
\end{tabular}
}
\caption{Accuracy (\%) of the LLaMA-3.1-8B model across a wide range of reasoning tasks: four \emph{Mathematical Reasoning} tasks (MultiArith, GSM8K, MATH, AQUA), four \emph{Commonsense Reasoning} tasks (CSQA, StrategyQA, BBH-Date Understanding, BBH-DisambiguationQA), and two \emph{Logical Reasoning} tasks (BBH-Logical Deduction, BBH-Web of Lies).
}
\label{tab:dataset_ablation}
\end{table*}

\paragraph{More Diverse Datasets}
Beyond the tasks featured in our main results, we expanded our evaluation to more varied datasets for a finer-grained ablation study with LLaMA-3.1-8B.
(a) \textit{AQUA}~\citep{ling2017program}: a multiple-choice set of arithmetic word problems with varying difficulty,
(b) \textit{CommonsenseQA}~\citep{talmor2019commonsenseqa}: a benchmark for commonsense reasoning,
(c) \textit{BigBench-Date Understanding} \citep{suzgun2023challenging}: which focuses on temporal and date-related reasoning,
(d) \textit{MATH}~\citep{hendrycks2021measuring}: high-school level math problems that extend beyond basic arithmetic,
(e) \textit{BBH-Web of Lies}~\citep{suzgun2023challenging}: tests related to logical reasoning in the presence of deceptive or incorrect statements.
Notably, GSM8K, MATH, and AQUA form an approximate difficulty hierarchy (GSM8K < MATH < AQUA) but exhibit similar improvement patterns under our approach. As shown in Table~\ref{tab:dataset_ablation}, our method provides substantial performance boosts over both zero-shot and zero-shot-CoT baselines across arithmetic, commonsense, and logical reasoning tasks. 
For instance, in the zero-shot setting, our approach improves MultiArith accuracy from 15.56\% to 50.56\% and GSM8K from 6.97\% to 34.57\%, while consistently outperforming baselines on other challenging datasets like MATH and BBH-Web of Lies. 
These gains underscore the versatility and effectiveness of our method in handling a wide range of reasoning tasks.

\begin{table}[!t]
\centering
\resizebox{\linewidth}{!}{%
\begin{tabular}{lcccc}
\toprule
 & \multicolumn{2}{c}{LLaMA-3.1-8B} & \multicolumn{2}{c}{Falcon3-10B-Base} \\
\cmidrule(lr){2-3}\cmidrule(lr){4-5}
 & Zero-shot & Zero-shot-CoT & Zero-shot & Zero-shot-CoT \\
\midrule\midrule
Baseline & 15.35\% & \textbf{29.13\%} & 43.70\% & 40.16\% \\
Ours     & \textbf{24.41\%} & 28.74\% & \textbf{57.09\%} & \textbf{51.57\%} \\
\bottomrule
\end{tabular}%
}
\caption{Performance (accuracy \%) on AQUA. ``Zero-shot'' and ``Zero-shot-CoT'' refer to the respective prompting strategies, with different model scales.}
\label{tab:model_scales}
\end{table}


\paragraph{Model Scales}
On the AQUA dataset, we found that at smaller scales (LLaMA-3.1-8B), the performance gains were rather minimal. However, once the model size increased to around 10B, the improvements became more evident. This indicates that no matter how much we refine the inherent capabilities, if the model itself does not have sufficient internal knowledge to begin with, the impact remains limited. In contrast, larger models tend to benefit more on challenging tasks, suggesting that once a model's internal capacity reaches a certain threshold, our method can further unlock its reasoning abilities.

For example, examining AQUA on smaller Llama models shows that the difference between Zero-shot (29.13\%) and Zero-shot CoT (28.74\%) is negligible. Yet, when moving to a 10B model, we see a clear boost in performance. This implies there is a point below which smaller models cannot substantially improve, even if we attempt to unlock their inherent reasoning capacity. Meanwhile, Falcon3-10B-Base~\citep{Falcon3} exhibits noticeable gains on AQUA, aligning with this trend.

\section{Related Works}
A variety of methods have been explored to enhance the reasoning capabilities of large language models (LLMs). These techniques can be broadly categorized into \emph{tuning-based}, \emph{prompt-based}, \emph{sampling-based}, and \emph{test-time} approaches. Below, we discuss how our decoding-based strategy relates to these prior efforts, especially in its ability to improve reasoning without extensive additional resources.

\paragraph{Tuning-Based Approaches}
Early research on boosting LLM reasoning focused on fine-tuning or distillation. For instance, instruction tuning \citep{wei2021finetuned} trains models on a broad set of tasks and instructions, improving zero-shot performance across multiple NLP benchmarks. System 2 distillation \citep{yu2023distilling} further refines model reasoning by transferring multi-step solutions from a larger teacher model. Although these methods can substantially enhance performance, they typically require large datasets and significant computational resources. Additionally, recent exploration into symbolic or chain-of-thought distillation \citep{li2024symbolicchainofthoughtdistillationsmall} still relies heavily on the quality of the larger model’s outputs. Our work diverges from these in that we tackle reasoning failures purely at inference time, without additional model training or data curation.

\paragraph{Prompt Engineering}
Prompt-based methods have garnered significant attention as a more resource-efficient alternative to fine-tuning. Chain-of-thought (CoT) \citep{wei2022chainofthought} structures responses into step-by-step explanations, while extensions such as Tree-of-Thoughts \citep{yao2023tree} and Graph-of-Thoughts \citep{besta2023graph} reorganize the reasoning flow into more complex hierarchies. Other lines of work employ iterative refinement, as in Self-Refine \citep{madaan2023selfrefine}, where the model iteratively critiques and updates its solution. Although these approaches can boost LLM reasoning, they often involve substantial prompt engineering or multiple rounds of inference, which can be costly in practice. By contrast, our method obviates heavy prompt tuning and avoids repeated sampling, opting instead for a lightweight, decoding-centric solution.

\paragraph{Sampling-Based Methods}
Approaches based on sampling seek to improve reliability of reasoning in LLMs by generating multiple candidate solutions. For example, self-consistency sampling \citep{wang2023selfconsistency} aggregates multiple outputs to select a consensus or majority-vote answer, while universal self-consistency \citep{chen2023universalselfconsistency} generalizes this concept to tasks where multiple lines of reasoning may converge on the correct outcome. These methods benefit from exploring diverse reasoning paths but come at the cost of multiple forward passes through the model. In many settings, this extra computational burden can be prohibitive. Our approach, in contrast, requires only one pass; we adaptively intervene during decoding to extend the chain of reasoning rather than generating multiple completions.

\paragraph{Test-Time Strategies}
More recent investigations have shifted attention to inference-time or post-processing methods that address reasoning shortcomings without altering the base model or training regimen. \citet{wang2023thoughts} introduces a penalty mechanism for abrupt shifts in reasoning flow, while \citet{snell2023scaling} proposes adaptive inference that strategically allocates resources for verification. Budget forcing \citep{muennighoff2025s1} constrains or terminates chain-of-thought decoding under a specified computational budget. Although these methods highlight the potential of inference-level interventions, many still rely on specialized heuristics or additional compute to manage extended reasoning steps.

By contrast, our work unlocks LLMs’ latent reasoning abilities \emph{without explicit prompting} or additional overhead. Specifically, we reduce \textit{immature reasoning} by inserting a brief injection phrase whenever the model risks early termination with the \texttt{<eos>} token. This simple yet powerful approach integrates seamlessly into a standard decoding pipeline and has proven effective in mitigating abrupt endings across diverse reasoning tasks.

\section{Conclusion}
In this work, we addressed the challenge of \emph{immature reasoning} in large language models (LLMs), wherein the model’s reasoning process terminates prematurely, leading to \emph{silence}, \emph{no reasoning}, or \emph{incomplete reasoning}. Our observations show that despite the substantial capacity of modern LLMs, this issue remains a major source of error in multi-step problem-solving.

To mitigate this, we proposed a novel decoding strategy. Rather than relying heavily on prompt engineering, our approach injects a designated “nudging” word whenever the model is likely to produce the \texttt{<eos>} token prematurely. This lightweight, test-time intervention encourages continued reasoning, acting as an implicit signal that mirrors human thinking process.

Extensive experiments across diverse reasoning tasks--including arithmetic, commonsense, and logical reasoning--demonstrate that our method substantially improves performance in both zero-shot and zero-shot-CoT settings. Moreover, it enhances reasoning stability, compensating for ineffective prompts and adapting to varying task complexities.

Overall, our proposed method, Adaptive Injection Decoding, represents a promising direction for unlocking the latent reasoning potential of LLMs while maintaining computational efficiency and ease of deployment. Future work could explore the dynamic selection of injected phrases based on task characteristics, the extension of our method to more complex generation tasks, or the integration with advanced sampling strategies (e.g., self-consistency, tree-of-thought decoding) to further enhance robustness and reliability.


\section*{Limitations}

While our method significantly enhances LLM performance, several limitations warrant attention. First, we tested models up to only 10 billion parameters, so the generalizability to extremely small models or much larger systems remains to be explored. Second, our approach relies on a simple injection phrase (``\texttt{Well}'') identified to perform robustly in experiments. Although it was proven effective on multiple tasks, variations in domain or language may require additional tuning of this trigger. Third, our experiments focus on reasoning benchmarks in English; extending to multilingual scenarios might reveal different patterns of \emph{immature reasoning} and alternative effective triggers.

\clearpage
\bibliographystyle{acl_natbib}
\bibliography{anthology,custom}
\clearpage

\appendix
\onecolumn

\section{Detailed Analysis of Injection Phrases} \label{sec:phrase_ablation}

In this section, we present a detailed study on the effectiveness of various phrase injections at test time, as summarized in Table~\ref{tab:token-ablation}. Our goal is to explore how small \emph{nudge} phrases can discourage premature termination and guide the model toward more extensive and coherent chains of thought. Consistent with our findings in Section~\ref{ablation}, we verify here that “\texttt{Well}” is the most consistently effective phrase among all options tested. This line of inquiry dovetails with prior research demonstrating that LLM decision-making can be strongly influenced by external prompts, including psychological or social pressures \citep{kim-etal-2024-will}.

\begin{table}[htb!]
\centering
\resizebox{0.6\textwidth}{!}{%
\begin{tabular}{c|c|c} \toprule
\multicolumn{1}{c}{Category} & \multicolumn{1}{c}{Phrase} & Accuracy \\ \hline\hline
\multirow{5}{*}{Machine Language} & \string<start of text> & 27.78\% \\
 & ------------ & 12.22\% \\
 & \textbackslash{}n & 14.44\% \\
 & \textbackslash{}t & 19.44\% \\
 & \# & 26.67\% \\ \hline
\multirow{6}{*}{Single Token} 
 & I & 32.22\%  \\
 & The & 33.33\% \\
 & Let & 38.33\%  \\
 & \textbf{Well} & \textbf{50.56\%} \\
 & Wait & 21.11\% \\
 & Step & 44.44\% \\ \hline
\multirow{7}{*}{Conjunction} 
 & And & 16.11\% \\
 & But & 17.78\% \\
 & Or & 12.78\%\\
 & So & 49.44\% \\
 & Therefore & 33.33\% \\
 & Because & 49.44\% \\
 & Alternatively & 20.56\% \\ \hline
\multirow{3}{*}{Conjunction Pool} 
 & Addition Pool & 26.11\% \\
 & Contrast Pool &  20.56\%\\
 & Mix Pool & 23.33\% \\ \hline
\multirow{2}{*}{Discourse Markers} 
 & I mean, & 27.22\% \\
 & You know, & 22.22\% \\ \hline
\multirow{2}{*}{\begin{tabular}[c]{@{}c@{}}Sentence \\(self-doubt)\end{tabular}} 
 & Am I doing alright? & 17.22\% \\
 & I might be wrong. & 33.33\%  \\ \hline
\multirow{2}{*}{\begin{tabular}[c]{@{}c@{}}Sentence\\(self-assurance)\end{tabular}} 
 & I can do it. & 17.78\% \\
 & I am doing quite well. & 13.38\% \\ \hline
\multirow{2}{*}{Someone else's voice} 
 & Keep reasoning. & 12.22\% \\
 & Think deeper. & 15.56\% \\ \bottomrule
\end{tabular}
}
\caption{Injection Phrase Ablation}
\label{tab:token-ablation}
\end{table}

\paragraph{Conclusion-Driven Phrases.} We observed that phrases prompting the model to continue toward a conclusion---for example, \emph{So}, \emph{Well}, and \emph{Because}---consistently yielded higher accuracies (e.g., 49.44\%, 50.56\%, and 49.44\%, respectively). By comparison, \emph{Therefore} performed relatively poorly (33.33\%), likely because it encourages the model to finalize its answer too abruptly. This pattern underscores why “\texttt{Well}” stood out in our main study: it nudges the model to continue reasoning without prematurely settling on a final answer.

\paragraph{Phrases Initiating Reasoning.} Certain phrases that denote a fresh start in thought, such as \emph{Let} and \emph{Step}, also contributed to improved outcomes (38.33\% and 44.44\%). Their effectiveness appears to stem from prompting the model to begin a new sequence of reasoning. However, these gains were still somewhat lower than those achieved by “\texttt{Well},” suggesting that “\texttt{Well}” provides a more neutral yet effective cue.

\paragraph{Continuity vs.\ Disruption.} Phrases like \emph{And}, which force the model to continue enumerating ideas, did not perform substantially better than the greedy baseline. Similarly, phrases that intentionally change the direction of thinking, such as \emph{But} or \emph{Or}, yielded low accuracy (17.78\% and 12.78\%), suggesting that abrupt shifts in reasoning can destabilize the model when it is already uncertain. Moreover, within the \emph{Conjunction Pool}, \emph{Addition Pool} achieves a higher accuracy (26.11\%) than both \emph{Mix Pool} (23.33\%) and \emph{Contrast Pool} (20.56\%). These findings align with the main study’s conclusion that gently reinforcing the existing reasoning path (e.g., via “\texttt{Well}”) is often more beneficial than forcing a sudden logical pivot.

\paragraph{Self-Doubt or Delay.} Although phrases like \emph{Wait} improve slightly on the baseline (21.11\% vs.\ 15.56\%), they do not show strong gains overall. We surmise that introducing hesitation without constructive guidance can heighten the model’s uncertainty. Full-sentence prompts expressing self-doubt (e.g., \emph{Am I doing alright?}) or self-assurance (\emph{I can do it.}) often led to immediate termination or repetitive output, thus hindering performance. Interestingly, the phrase \emph{I might be wrong.} reached a markedly higher accuracy (33.33\%), echoing the discussion in Section~\ref{ablation} that acknowledging potential error can sometimes encourage more careful reasoning.

\paragraph{Machine Language and Formatting.} Interestingly, the use of machine-like phrases (e.g., \texttt{\#} or \texttt{\string<start of text>}) reached accuracies near or above 25\%, indicating that inducing a formatting or structural shift can be beneficial. We suspect this formatting cue encourages more systematic thinking and helps avert premature termination. Similarly, inserting \verb|\t| improved upon the greedy baseline (19.44\% vs.\ 15.56\%), possibly by delaying the immediate production of \texttt{<eos>}. However, more verbose sentence-level injections (e.g., \emph{Keep reasoning.}, \emph{Think deeper.}) tended to cause the model to either terminate immediately or restate these prompts, which hindered performance.

\paragraph{Discussion.} Overall, these results corroborate our main finding that “\texttt{Well}” is the most consistently effective injection phrase. Phrases that naturally sustain or refocus the chain of thought---without dictating a final answer---tend to achieve higher accuracy. “\texttt{Well}” exemplifies a subtle yet impactful nudge, fostering continued reasoning while preserving flexibility across various problem types. Future investigations might consider how sentence-level injections (e.g., explicit self-doubt such as \emph{I might be wrong.}) would scale in larger LLMs, where nuanced disclaimers could further modulate reasoning under stress or social pressures \citep{kim-etal-2024-will}.

\subsection{Illustrative Examples of Phrase Injections}
\label{subsec:illustrative_examples}

Below, we provide concrete examples demonstrating how specific injection words or phrases can influence the model's responses. These examples illuminate common behaviors (e.g., \textbf{silence}, \textbf{no reasoning}, \textbf{incomplete reasoning}) that certain phrases can mitigate—\emph{or} in some cases exacerbate. By examining both improvements and adverse outcomes, we show that while injection phrases often help sustain or refocus the reasoning process, they may also inadvertently introduce tangential or repetitive outputs if misapplied. 

\newpage
\paragraph{Silence.}
When the model (greedy) immediately outputs \texttt{<eos>}, injecting a phrase like \emph{Because} can encourage it to elaborate:

\begin{table}[h]
\centering
\caption{Zero-Shot Case. The model outputs <eos> token on the first generation, showing \textbf{Silence}.}
\begin{tcolorbox}[width=\textwidth, colback=white, colframe=black, boxsep=5pt, arc=2pt]
\textbf{Q:} For Halloween Debby and her sister combined the candy they received. Debby had 32 pieces of candy while her sister had 42. If they ate 35 pieces the first night, how many pieces do they have left? 

\textbf{A:} <eos> (\textbf{Silence})
\end{tcolorbox}
\end{table}

\begin{table}[h]
\centering
\caption{Injecting \emph{Because} helps the model overcome silence, ultimately producing the answer.}
\begin{tcolorbox}[width=\textwidth, colback=white, colframe=black, boxsep=5pt, arc=2pt]
\textbf{Q:} For Halloween Debby and her sister combined the candy they received. Debby had 32 pieces of candy while her sister had 42. If they ate 35 pieces the first night, how many pieces do they have left? 

\textbf{A:} \emph{Because} the problem is asking for the total number of pieces left, we need to find the total number of pieces they had. To do this, we need to add the number of pieces each had. 32 + 42 = 74. Now that we know the total number of pieces they had, we can subtract the number of pieces they ate to find the number of pieces left. 74 - 35 = 39. Therefore, they have 39 pieces of candy left.
\end{tcolorbox}
\end{table}

\paragraph{No Reasoning.}
Similarly, \emph{So} can redirect the model from meaningless enumeration to a more coherent answer:

\begin{table}[H]
\centering
\caption{Zero-Shot Case. The model repeats meaningless enumeration until reaching the max generation length.}
\begin{tcolorbox}[width=\textwidth, colback=white, colframe=black, boxsep=5pt, arc=2pt]
\textbf{Q:} Paige had 11 songs on her mp3 player. If she deleted 9 old songs from it and then added 8 new songs, how many songs does she have on her mp3 player?

\textbf{A:} 1.  2.  3.  4.  5.  6.  7.  8.  9.  10.  11.  12.  13.  14.  15.  16.  17.  18.  19.  20.  21.  22.  23.  24.  25.  26.  27.  28.  29.  30.  31.  32.  33.  34.  35.  36.  37.  38.  39.  40.  41.  42.  43.  44.  45.  46.  47.  48.  49.  50.  51.  52.  53.  54.  55.  56.  57.  58.  59.  60.  61.  62.  63.  64.  65.  66.  67.  68.  69.  70.  71.  72.  73.  74.  75.  76.  77.  78.  79.  80.  81.  82.  83.  84.  85.  86.  87.  88.  89.  90.  91.  92.  93.  94.  95.  96.  97.  98.  99.  100.  101.  102.  103.  104.  105.  106.  107.  108.  109.  110.  111.  112.  113.  114.  115.  116.  117.  118.  119.  120.  121.  122.  123.  124.  125. 
\end{tcolorbox}
\end{table}

\begin{table}[H]
\centering
\caption{Injecting \emph{So} helps the model avoid repetitive output and focus on problem solving.}
\begin{tcolorbox}[width=\textwidth, colback=white, colframe=black, boxsep=5pt, arc=2pt]
\textbf{Q:} Paige had 11 songs on her mp3 player. If she deleted 9 old songs from it and then added 8 new songs, how many songs does she have on her mp3 player?

\textbf{A:}  1.  2.\emph{So}, Paige has 10 songs on her mp3 player.  
\end{tcolorbox}
\end{table}
\newpage
\paragraph{Incomplete Reasoning.}
Injecting \emph{Well} prompts the model to provide step-by-step reasoning instead of a rushed or partial conclusion:

\begin{table}[H]
\centering
\caption{The model provides the correct choice but without showing any reasoning process.}
\begin{tcolorbox}[width=\textwidth, colback=white, colframe=black, boxsep=5pt, arc=2pt]
\textbf{Q:} Bianca and her friends were recycling paper for their class. For every 3 pounds they recycled they earned one point. If Bianca recycled 24 pounds and her friends recycled 3 pounds, how many points did they earn?

\textbf{A:}   8  9  10  11
A.  7  B.  8  C.  9  D.  10
Answer: C
\end{tcolorbox}
\end{table}

\begin{table}[H]
\centering
\caption{Injecting \emph{Well} helps the model complete its reasoning.}
\begin{tcolorbox}[width=\textwidth, colback=white, colframe=black, boxsep=5pt, arc=2pt]
\textbf{Q:} Bianca and her friends were recycling paper for their class. For every 3 pounds they recycled they earned one point. If Bianca recycled 24 pounds and her friends recycled 3 pounds, how many points did they earn?

\textbf{A:}   8  9  10  11
A.  7  B.  8  C.  9  D.  10
Answer: C \emph{Well}, Bianca and her friends recycled 24 pounds and 3 pounds, respectively. So, the total pounds they recycled is 24 + 3 = 27 pounds. For every 3 pounds they recycled they earned one point. So, the total points they earned is 27 / 3 = 9 points.
\end{tcolorbox}
\end{table}

\paragraph{Therefore.}
While \emph{Therefore} might sometimes help finalize a correct answer, we also observed instances where it led to premature or abrupt conclusions:

\begin{table}[H]
\centering
\caption{A case where injecting \emph{Therefore} helps the model arrive at the correct answer.}
\begin{tcolorbox}[width=\textwidth, colback=white, colframe=black, boxsep=5pt, arc=2pt]
\textbf{Q:} Gwen was organizing her book case making sure each of the shelves had exactly 4 books on it. If she had 5 shelves of mystery books and 3 shelves of picture books, how many books did she have total? 

\textbf{A:} Gwen was organizing her book case making sure each of the shelves had exactly 4 books on it. If she had 5 shelves of mystery books and 3 shelves of picture books, how many books did she have total? \textit{(The model initially repeats the question.)} \emph{Therefore}, the total number of books is 5*4 + 3*4 = 32 books.
\end{tcolorbox}
\end{table}

\begin{table}[H]
\centering
\caption{A case where injecting \emph{Therefore} causes the model to produce an abrupt, incorrect conclusion.}
\begin{tcolorbox}[width=\textwidth, colback=white, colframe=black, boxsep=5pt, arc=2pt]
\textbf{Q:} Faye had 34 coloring books. If she gave away 3 of them, but then bought 48 more, how many would she have total? 

\textbf{A:} A. 79 B. 81 C. 82 D. 83 E. 84 \emph{Therefore}, the correct answer is C. \textit{(The real answer is 79.)}
\end{tcolorbox}
\end{table}
\newpage
\paragraph{Let, Step.}
Phrases like \emph{Let} or \emph{Step} can initialize a new reasoning chain, though they may also diverge from the original chain:

\begin{table}[H]
\centering
\caption{A case where injecting \emph{Let} helps, by initializing a fresh reasoning chain.}
\begin{tcolorbox}[width=\textwidth, colback=white, colframe=black, boxsep=5pt, arc=2pt]
\textbf{Q:} For a birthday party Jerry bought 41 regular sodas and 22 diet sodas. If his fridge would only hold 9 on each shelf, how many shelves would he fill up? 

\textbf{A:} \emph{Let} x be the number of shelves. 9x = 41 + 22 9x = 63 x = 7 Jerry would need 7 shelves to hold all the sodas.
\end{tcolorbox}
\end{table}

\begin{table}[H]
\centering
\caption{A case where injecting \emph{Let} does not help, as it starts a completely new reasoning chain.}
\begin{tcolorbox}[width=\textwidth, colback=white, colframe=black, boxsep=5pt, arc=2pt]
\textbf{Q:} In a video game, each enemy defeated gives you 9 points. If a level has 11 enemies total and you destroy all but 3 of them, how many points would you earn? 

\textbf{A:} 9 x 8 = 72 points.\emph{Let}'s say you have 3 enemies left and you destroy them all. 9 x 3 = 27 points. 72 + 27 = 99 points. \textit{(The correct answer is 72.)}
\end{tcolorbox}
\end{table}

\paragraph{And, But.}
Conjunctions such as \emph{And} or \emph{But} can cause the model to diverge into tangential queries or endless hypotheticals:

\begin{table}[H]
\centering
\caption{A case where injecting \emph{And} causes the model to introduce tangential queries.}
\begin{tcolorbox}[width=\textwidth, colback=white, colframe=black, boxsep=5pt, arc=2pt]
\textbf{Q:} A company invited 47 people to a luncheon, but 7 of them didn't show up. If the tables they had held 5 people each, how many tables do they need? 

\textbf{A:} \emph{And} if the tables they had held 6 people each, how many tables do they need? 
\end{tcolorbox}
\end{table}

\begin{table}[H]
\centering
\caption{A case where injecting \emph{But} causes the model to spiral into endless scenarios.}
\begin{tcolorbox}[width=\textwidth, colback=white, colframe=black, boxsep=5pt, arc=2pt]
\textbf{Q:} In a video game, each enemy defeated gives you 9 points. If a level has 11 enemies total and you destroy all but 3 of them, how many points would you earn?  

\textbf{A:} 9 x 8 = 72 points. \emph{But} what if you only destroyed 2 of the enemies? 9 x 2 = 18 points. What if you destroyed 4 of the enemies? 9 x 4 = 36 points. What if you destroyed 5 of the enemies? 9 x 5 = 45 points. What if you destroyed 6 of the enemies? 9 x 6 = 54 points. What if you destroyed 7 of the enemies? 9 x 7 = 63 points. What if you destroyed 8 of the enemies? 9 x 8 = 72 points. What if you destroyed 9 of the enemies? 9 x 9 = 81 points. What if you destroyed 10 of the enemies? 9 x 10 = 90 points. What if you destroyed 11 of the enemies? 9 x 11 = 99 points. What if you destroyed 12 of the enemies? 9 x 12 = 108 points. What if you destroyed 13 of the enemies? 9 x 13 = 117 points. What if you destroyed 14 of the enemies? 9 x 14 = 126 points. What if you destroyed 15 of the enemies? 9 x 15 = 135 points. What if you destroyed 16 of the enemies? 9 x 16 = 144 points. What if you destroyed 17 of the enemies? 9 x 17 = 153 points. What if you destroyed 18 of the enemies? 9 x 18 = 162 points. What if you destroyed 19 of the enemies? 9 x 19 = 171 points. What if you (...keeps making endless scenarios)

\end{tcolorbox}
\end{table}

\newpage

\paragraph{I might be wrong.}
Conveying self-doubt can allow the model to proceed more cautiously and produce a correct answer, rather than halting:

\begin{table}[H]
\centering
\caption{A case where injecting \emph{I might be wrong.} helps the model proceed and produce the correct answer.}
\begin{tcolorbox}[width=\textwidth, colback=white, colframe=black, boxsep=5pt, arc=2pt]
\textbf{Q:} For Halloween Emily received 5 pieces of candy from neighbors and 13 pieces from her older sister. If she only ate 9 pieces a day, how long would the candy last her?

\textbf{A:} \emph{I might be wrong.} I think it would last 2 days.
\end{tcolorbox}
\end{table}

\paragraph{I can do it., I am doing quite well.}
Self-affirming phrases can lead to overconfidence or repetitive affirmations without solving the problem:

\begin{table}[H]
\centering
\caption{A case where injecting \emph{I can do it.} causes the model to be overconfident, offering no real solution.}
\begin{tcolorbox}[width=\textwidth, colback=white, colframe=black, boxsep=5pt, arc=2pt]
\textbf{Q:} A store had 40 oranges in a bin. If they threw away 25 of the old ones and put 21 new ones in the bin how many would be in the bin? 

\textbf{A:} \emph{I can do it.} I just don't know how to explain it.
\end{tcolorbox}
\end{table}

\begin{table}[H]
\centering
\caption{A case where injecting \emph{I am doing quite well.} causes the model to loop repetitive affirmations.}
\begin{tcolorbox}[width=\textwidth, colback=white, colframe=black, boxsep=5pt, arc=2pt]
\textbf{Q:} Faye had 34 coloring books. If she gave away 3 of them, but then bought 48 more, how many would she have total? 

\textbf{A:} A. 79 B. 81 C. 82 D. 83 E. 84\emph{I am doing quite well.} I am in the 90th percentile. I am in the 90th percentile. I am in the 90th percentile. I am in the 90th percentile. I am in the 90th percentile. I am in the 90th percentile. I am in the 90th percentile. I am in the 90th percentile. I am in the 90th percentile. I am in the 90th percentile. I am in the 90th percentile. I am in the 90th percentile. I am in the 90th percentile. I am in the 90th percentile. I am in the 90th percentile. I am in the 90th percentile. I am in the 90th percentile. I am in the 90th percentile. I am in the 90th percentile. I am in the 90th percentile. I am in the 90th percentile. I am in the 90th percentile. I am in the 90th percentile. I am in the 90th percentile. I am in the 90th percentile. I am in the 90th percentile. I am in the 90th percentile. I am in the 90th percentile. I am in the 90th percentile. I am in the 90th percentile. I am in the 90th percentile. I am in the 90th percentile. I am in the 90th percentile. I am in the 90th percentile. I am in the 90th percentile. I am in the 90th percentile. I am in the 90th percentile. I am in the 90th percentile. I am in the 90th percentile. I am in the 90th percentile. I am in the 90th percentile. I am in the 90th percentile. I am in the 90th percentile. I am in the 90th percentile. I am in the 90th percentile. I am in the 90th percentile. I am in the 90th percentile. I am in the 90th percentile. I am in the 90th percentile. I am in the 90th percentile. I am in the 90th percentile. I am in the 90th percentile. I am in the 90th percentile. I am
\end{tcolorbox}
\end{table}

\newpage

\section{Conjunction Pools}
\label{sec:conjunction_pools}

As noted in Section~\ref{sec:phrase_ablation} (and shown in Table~\ref{tab:token-ablation}), conjunction-based phrases exhibit notable differences depending on whether they reinforce existing reasoning or introduce a sharp directional shift. To introduce controlled variability rather than relying on a fixed injection phrase, we constructed a conjunction pool consisting of semantically similar phrases. During inference, the injection phrase is randomly sampled from this pool, allowing for slight variation while preserving its intended effect. Table~\ref{tab:pool_conjunctions} presents the complete lists used in our experiments.

\begin{table}[h]
    \centering
    \caption{Conjunction pools used in our experiments.}
    \label{tab:pool_conjunctions}
    \begin{tabular}{l p{.55\linewidth}}
        \toprule
        Pool & Conjunctions \\
        \hline\hline
        \textit{ADDITION\_POOL} & 
        \{"and", "so", "therefore", "then", "thus", "or", "in addition", "furthermore"\} \\
        \textit{CONTRAST\_POOL} & 
        \{"however", "but", "on the other hand", "yet", "in contrast", "nevertheless", "unlike", "instead", "conversely"\} \\
        \textit{MIX\_POOL} & \textit{ADDITION\_POOL} + \textit{CONTRAST\_POOL} \\
        \bottomrule
    \end{tabular}
\end{table}

Specifically, we define separate pools for \textit{addition} conjunctions, \textit{contrast} conjunctions, and a \textit{mixed} set that combines both. For each instance, an injection phrase is randomly drawn from the relevant pool: the \textit{ADDITION\_POOL} for reinforcing prior reasoning, the \textit{CONTRAST\_POOL} for shifting direction, and the \textit{MIX\_POOL} for a broader range of transitions. This randomized selection enables a more flexible strategy while allowing us to systematically analyze the impact of different conjunction types on model performance.


\section{Evaluation Prompt and Human Alignment}
We use \texttt{o1-mini} via OpenAI’s chat completion API (version 2024-09-12) \citep{openai2024o1mini} to assess correctness in our experiments. In scenarios where LLMs produce freely generated responses rather than strictly formatted answers, extracting definitive outcomes can be challenging. Consequently, LLM-based evaluation becomes a practical solution. Among the models we tested, \texttt{o1-mini} showed the most robust performance and closely matched human judgments, even under these open-ended conditions.

As \texttt{o1-mini} demonstrates advanced language understanding, we specifically designed a detailed set of grading criteria to minimize misclassification and ensure consistent evaluation outcomes. Our criteria aim to address various answer formats and potential pitfalls. In particular, the rules instruct \texttt{o1-mini} to:
\begin{enumerate}
    \item Compare only the final stated answer with the ground truth, ignoring any intermediate reasoning provided by the large language model (LLM).
    \item Verify that the final answer choice aligns with the LLM’s self-generated options (if any), marking mismatched selections as incorrect.
    \item Disregard auxiliary examples or sub-problems introduced by the LLM as part of its reasoning, focusing strictly on the main query and final response.
    \item Consider minor variations in phrasing valid, as long as the intended meaning remains correct.
    \item For True/False questions, infer correctness from the overall meaning of the response (e.g., a statement affirming or denying a proposition) when an explicit “true” or “false” label is not provided.
\end{enumerate}

\label{sec:evaluation_prompt}
\newpage
\begin{tcolorbox}[
  colback=blue!5!white,
  colframe=blue!75!black,
  title=Evaluation Prompt,
  breakable  
]
\small          
\begin{verbatim}
========================================================
GOAL
========================================================
You are given a question, its correct answer (ground truth), and an LLM’s response. 
Your task is to determine whether the LLM’s final answer matches the correct answer.

========================================================
RETURN FORMAT
========================================================
Respond with either “correct” or “incorrect” only.

========================================================
EVALUATION RULES
========================================================
1. Always judge based on the final answer given by the LLM.  
   - If the LLM provides reasoning before arriving at a final answer, ignore intermediate steps 
     and only compare the last stated answer with the ground truth.  
   - Example:
     ```
     LLM’s Response: Well, the answer is 24 hours. But how did we get there? 
     I know that the total amount of money he earned is 7 times the amount of money 
     he earned from each customer. So, 7(3) = 21.
     ```
     The final answer here is **21**, so compare it with the ground truth.

2. If the LLM selects an incorrect choice from predefined options generated by itself, 
   mark it incorrect—even if it derives the correct answer in its reasoning.  
   - Example:
     ```
     LLM’s Response: (A) 11 (B) 13 (C) 15
     Well, I know that 50 - 40 = 10. So she had 10 eggs.
     The final answer is (A).
     ```
     The correct answer is **10**, but the LLM’s selected choice **(A) 11** is incorrect.

3. Ignore example problems the LLM generates during reasoning.  
   - If the LLM solves auxiliary example problems generated by itself before answering the 
     given question, disregard those and judge based only on its answer to the main question.  
   - Example:
     ```
     Question: James decides to run 3 sprints 3 times a week. He runs 60 meters each sprint. 
     How many total meters does he run a week?
     Answer: 540

     LLM’s Response:  
     The distributive property states that...  
     #### Example A  
     Solve for x.  
     3(x+2) = 15 → x = 3  
     #### Example B  
     2(x-3) = 10 → x = 8  
     #### Example C  
     4(x+5) = 20 → x = 0  

     Concept Problem Revisited  
     James decides to run 3 sprints 3 times a week. He runs 60 meters each sprint.  
     3 × 3 = 9  
     9 × 60 = 540  

     James runs 540 meters a week.
     ```
     The final answer is **540**, which should be compared to the ground truth.

4. Handle answer format variations properly.

  - 4.1. If the LLM’s answer matches a choice’s meaning but not its letter label, mark it correct.  
    - Example:
      ```
      Question:
      What is likely to satisfy someone's curiosity? (A) hear news (B) read book 
      (C) see favorite show (D) comedy show (E) go somewhere

      Answer: A

      LLM’s Response: The answer is hearing news.
      ```
      Since "hearing news" corresponds to choice (A), this is correct.

  - 4.2. For multiple-choice questions where the LLM outputs only the letter choice, 
         ensure it matches the meaning of the correct answer.  
    - Example:
      ```
      Answer: (B) read book 
      LLM’s Response: B
      ```
      This is correct since (B) corresponds to "read book".

  - 4.3. Compare meaning, allowing for slight rewording as long as the answer’s intent is the same.

  - 4.4. For True/False Questions, if the LLM does not explicitly state “true” or “false,” 
         infer its judgment based on the response’s meaning before comparison. 
    - Example:
      ```
      Question: Does the moon have its own light? Answer: False

      LLM’s Response: The moon does not generate its own light; it only reflects sunlight.
      ```
      This should be marked correct, as the meaning aligns with "false".

  Question: {question}
  Answer: {answer}
  LLM's Response: {llm_answer}

  Is the predicted answer correct?
\end{verbatim}
\end{tcolorbox}

\begin{table}[H]
\centering
\begin{tabular}{l|ccccc}
\toprule
Dataset & Multiarith & GSM8K & StrategyQA & DisambiguationQA & Logical Deduction \\
\midrule\midrule
gpt-4o-mini & 29.44\% & 24.56\% & 6.00\% & 15.00\% & 38.00\% \\
o1-mini      & 15.56\% & 6.97\% & 20.00\% & 40.00\% & 39.00\% \\
Human        & 11.11\% & 2.65\% & 25.00\% & 49.00\% & 38.00\% \\
\bottomrule
\end{tabular}
\caption{Comparison of LLM-based evaluations (using \texttt{o1-mini} and \texttt{gpt-4o-mini}) with human evaluation. All results were obtained via zero-shot generation using LLaMA-3.1-8B.}
\label{tab:alignment}
\end{table}

\clearpage
\section{Examples of Immature Reasoning }
\label{sec:premature-termination}

Immature reasoning refers to instances where LLMs fail to complete the Chain of Thought (CoT) reasoning process. To better understand the nature of these failures, we categorize them into four types and provide real examples demonstrating each failure pattern. Each failure type is presented with a clear example, including the input prompt, the model's incorrect response.

\subsection{\textit{Silence}: Failure to Generate Any Response}
\label{sec:silence}
\textit{Silence} occurs when the model fails to produce \emph{any} output in response to a given question. Instead of providing even a partial solution or a refusal, the model simply emits no tokens. As shown in our main analysis, this pattern accounted for 58.05\% of incorrect responses in LLaMA 3.1-8B's GSM8K outputs.

\begin{table}[H]
\centering
\caption{LLaMA-3.1-8B (on GSM8K) terminates with <eos> in 58.05\% of problems, failing to generate a complete response.}
\begin{tcolorbox}[width=\textwidth, colback=white, colframe=black, boxsep=5pt, arc=2pt]
\textbf{Q:} Janet’s ducks lay 16 eggs per day. She eats three for breakfast every morning and bakes muffins for her friends every day with four. She sells the remainder at the farmers' market daily for 2 per fresh duck egg. How much in dollars does she make every day at the farmers' market?

\textbf{A:}  <eos>
\end{tcolorbox}
\end{table}

\subsection{\textit{No Reasoning}: Generating a Response Without Reasoning}
\label{sec:no-reasoning}
\textit{No Reasoning} refers to answers that do not exhibit \emph{any} chain-of-thought or logical progression. In these cases, the model usually outputs repeated phrases, irrelevant content (e.g., HTML tags or unrelated numeric sequences), or a simple restatement of the question without meaningful analysis.

\begin{table}[H]
\centering
\caption{Gemma 7B (on MultiArith) repeating the given question instead of solving it.}
\begin{tcolorbox}[width=\textwidth, colback=white, colframe=black, boxsep=5pt, arc=2pt]
\textbf{Q:} The school cafeteria had 23 apples. If they used 20 to make lunch for the students and then bought 6 more, how many apples would they have?

\textbf{A:} The school cafeteria had 23 apples. If they used 20 to make lunch for the students and then bought 6 more, how many apples would they have? The school cafeteria had 23 apples. If they used 20 to make lunch for the students and then bought 6 more, how many apples would they have?  The school cafeteria had 23 apples. If they used 20 to make lunch for the students and then bought 6 more, how many apples would they have? The school cafeteria had 23 apples. If they used 20 to make lunch for the students and then bought 6 more, how many apples would they have? The school cafeteria had 23 apples. If they used 20 to make lunch for the students and then bought 6 more, how many apples would they have? The school cafeteria had 23 apples. If they used 20 to make lunch for the students and then bought 6 more, how many apples would they have? The school cafeteria had 23 apples. If they used 20 to make lunch for the students and then bought 6 more, how many apples would they have? The school cafeteria had 23 apples. If they used 20 to make lunch for the students and then bought 6 more, how many apples would they have? The school cafeteria had 23 apples. If they used 20 to make lunch for the students and then bought 6 more, how many apples would they have? The school cafeteria had 23 apples. If they used 20 to make lunch for the students and then bought 6 more, how many apples would they have? The school cafeteria had 23 apples. If they used 20 to make lunch for the students and then bought 6 more, how many apples would they have? The school cafeteria had 23 apples. If they used 20 to make lunch for the students and then bought 6 more, how many apples
\end{tcolorbox}
\end{table}

\vspace{-2em}

\begin{table}[H]
\centering
\caption{Mistral-7B (on CommonsenseQA) produces unrelated grammar questions in 42.49\% of its responses, as exemplified below.}
\begin{tcolorbox}[width=\textwidth, colback=white, colframe=black, boxsep=5pt, arc=2pt]
\textbf{Q:} What do people aim to do at work?
    (A) complete job
    (B) learn from each other
    (C) kill animals
    (D) wear hats
    (E) talk to each other

\textbf{A:} Random Topics: Adjectives and AdverbsAdjectives and AdverbsAdjectives and Adverbs  Next quiz: Grammar  The \_\_\_\_\_\_\_\_ of the sentence is the subject.  A. verb B. noun C. adjective D. adverb
\end{tcolorbox}
\end{table}

\begin{table}[H]
\centering
\caption{Mistral 7B (on MultiArith) outputs irrelevant numeric sequences and terminates generation.}
\begin{tcolorbox}[width=\textwidth, colback=white, colframe=black, boxsep=5pt, arc=2pt]
\textbf{Q:} Ned had to wash 9 short sleeve shirts and 21 long sleeve shirts before school. If he had only washed 29 of them by the time school started, how many did he not wash?

\textbf{A:} 1. 2. 3. 4. 5. 6. 7. 8. 9. 10. 11. 12. 13. 14. 15. 16. 17. 18. 19. 20. 21. 22. 23. 24. 25. 26. 27. 28. 29. 30. 31. 32. 33. 34. 35. 36. 37. 38. 39. 40. 41. 42. 43. 44. 45. 46. 47. 48. 49. 50. 51. 52. 53. 54. 55. 56. 57. 58. 59. 60. 61. 62. 63. 64. 65. 66. 67. 68. 69. 70. 71. 72. 73. 74. 75. 76. 77. 78. 79. 80. 81. 82. 83. 84. 85. 86. 87. 88. 89. 90. 91. 92. 93. 94. 95. 96. 97. 98. 99. 100. 101. 102. 103. 104. 105. 106. 107. 108. 109. 110. 111. 112. 113. 114. 115. 116. 117. 118. 119. 120. 121. 12
\end{tcolorbox}
\end{table}

\begin{table}[H]
\centering
\caption{Llama3 8B (on CommonsenseQA) producing additional, unnecessary choices in multiple-choice scenarios.}
\begin{tcolorbox}[width=\textwidth, colback=white, colframe=black, boxsep=5pt, arc=2pt]
\textbf{Q:} When drinking booze what can you do to stay busy?
    (A) reach tentative agreement
    (B) stay in bed
    (C) stop bicycle
    (D) examine thing
    (E) suicide

\textbf{A:} (F) stay busy (G) stay sober (H) stay in shape 

Answer: F 

Explanation: When drinking booze what can you do to stay busy? When drinking booze what can you do to stay busy?
\end{tcolorbox}
\end{table}

\begin{table}[H]
\centering
\caption{Gemma 7B (on MultiArith) arbitrarily generating answer choices even for non-multiple-choice questions.}
\begin{tcolorbox}[width=\textwidth, colback=white, colframe=black, boxsep=5pt, arc=2pt]
\textbf{Q:} Luke was helping the cafeteria workers pick up lunch trays, but he could only carry 4 trays at a time. If he had to pick up 20 trays from one table and 16 trays from another, how many trips will he make?

\textbf{A:} A. 4 \\  \\ B. 5 \\  \\ C. 6 \\  \\ D. 7 \\  \\ E. 8 \\  \\ Show more \\ Step 1 \\ 1 of 2 \\  \\ Determine the number of trips Luke will make. \\  \\ \begin{gather*} 20 \div 4 = 5 \text{ trips} \end{gather*} \\  \\ Result \\ 2 of 2 \\  \\ B. 5 

Answer: F 

Explanation: When drinking booze what can you do to stay busy? When drinking booze what can you do to stay busy?
\end{tcolorbox}
\end{table}

\begin{table}[H]
\centering
\caption{Gemma-7B (on Logical Deduction) generating an answer first before reasoning through it, but the subsequent solution lacks meaningful reasoning.}
\begin{tcolorbox}[width=\textwidth, colback=white, colframe=black, boxsep=5pt, arc=2pt]
\textbf{Q:} The following paragraphs each describe a set of five objects arranged in a fixed order. The statements are logically consistent within each paragraph. On a shelf, there are five books: a brown book, a yellow book, an orange book, a green book, and a gray book. The green book is to the left of the gray book. The brown book is the third from the left. The gray book is the second from the right. The yellow book is to the left of the green book.

Options:

(A) The brown book is the third from the left 

(B) The yellow book is the third from the left

(C) The orange book is the third from the left 

(D) The green book is the third from the left 

(E) The gray book is the third from the left

\textbf{A:} 

(C) The orange book is the third from the left\\ \\ 
                    Solution: \\ \\ 
                    The green book is to the left of the gray book.\\ 
                    The brown book is the third from the left.\\ 
                    The gray book is the second from the right.\\ 
                    The yellow book is to the left of the green book.\\ \\ 
                    The green book is to the left of the gray book.\\ 
                    The brown book is the third from the left.\\ 
                    The gray book is the second from the right.\\ 
                    The yellow book is to the left of the green book.\\ \\ 
                    The green book is to the left of the gray book.\\ 
                    The brown book is the third from the left.\\ 
                    The gray book is the second from the right.\\ 
                    The yellow book is to the left of the green book.\\ \\ 
                    The green book is to the left of the gray book.\\ 
                    The brown book is the third from the left.\\ 
                    The gray book is the second from the right.\\ 
                    The yellow book is to the left of the green book.
\end{tcolorbox}
\end{table}

\begin{table}[H]
\centering
\caption{Mistral-7B (on MultiArith) frequently generates "\# Guidance" and "\# Example", mimicking a few-shot style, which reflects the model’s learned pattern for mathematical problem-solving.}
\begin{tcolorbox}[width=\textwidth, colback=white, colframe=black, boxsep=5pt, arc=2pt]
\textbf{Q:} The school cafeteria had 23 apples. If they used 20 to make lunch for the students and then bought 6 more, how many apples would they have?

\textbf{A:} 26 \\ 
\#\#\# Guidance \\
The distributive property is a property of multiplication that allows us to distribute a factor over a sum or difference. \\
\[
\begin{aligned}
3(a+b) &= 3a+3b \\
5(a-b) &= 5a-5b
\end{aligned}
\]
The distributive property is used to simplify expressions. \\

\#\#\#\# Example A \\
Simplify: \$5(a+b)\$ \\
Solution: Use the distributive property to distribute the 5 over the sum. \\
\[
\begin{aligned}
5(a+b) &= 5a+5b \\
       &= 5a+5b
\end{aligned}
\]

\#\#\#\# Example B \\
Simplify: \$3(a-b)\$ \\
Solution: Use the distributive property to distribute the 3 over the difference. \\
\[
\begin{aligned}
3(a-b) &= 3a-3b \\
       &= 3a-3b
\end{aligned}
\]
\\

\#\#\#\# Example C \\
Simplify: \$2(a+b+c)\$ \\
Solution: Use the distributive property to distribute the 2 over the sum. \\
\[
\begin{aligned}
2(a+b+c) &= 2a+2b+2c \\
         &= 2a+2b+2c
\end{aligned}
\]
\\

\#\#\#\# Concept Problem Revisited \\
The school cafeteria had 23 apples. If they used 20 to make lunch for the students and then bought 6 more, how many apples would they have? \\
The cafeteria had 23 apples. They used 20 to make lunch for the students. They bought 6 more. \\
\[
\begin{aligned}
23-20 &= 3 \\
20+3  &= 23
\end{aligned}
\]

The cafeteria had 23 apples. They used 20 to make lunch for the students. They bought 6 more. \\
\[
\begin{aligned}
23-20 &= 3 \\
20+3  &= 23
\end{aligned}
\]

The cafeteria had 23

\end{tcolorbox}
\end{table}

\clearpage

\subsection{\textit{Incomplete Reasoning}: Resulting in an Incorrect or Partial Answer}
\label{sec:incomplete-reasoning}
\textit{Incomplete Reasoning} encompasses scenarios where the model initiates a chain of thought but stops prematurely or changes topic before reaching the correct conclusion. This phenomenon accounted for 5.45\% of incorrect GSM8K responses in LLaMA 3.1-8B.

\begin{table}[H]
\centering
\caption{LLaMA-3 8B (on MultiArith) failing to derive the final answer by prematurely stopping the Chain-of-Thought (CoT) process. The correct answer is 80, but the model halts before completing the final step of summing 32 and 48.}
\begin{tcolorbox}[width=\textwidth, colback=white, colframe=black, boxsep=5pt, arc=2pt]
\textbf{Q:} Zoe was unboxing some of her old winter clothes. She found 8 boxes of clothing and inside each box there were 4 scarves and 6 mittens. How many pieces of winter clothing did Zoe have total?

\textbf{A:} 1. 8 scarves and 6 mittens 2. 8 scarves and 24 mittens 3. 32 scarves and 24 mittens 4. 32 scarves and 48 mittens \\ Zoe was unboxing some of her old winter clothes. She found 8 boxes of clothing and inside each box there were 4 scarves and 6 mittens. How many pieces of winter clothing did Zoe have total? 1. 8 scarves and 6 mittens 2. 8 scarves and 24 mittens 3. 32 scarves and 24 mittens 4. 32 scarves and 48 mittens

\end{tcolorbox}
\end{table}

\begin{table}[t]
\centering
\caption{Mistral 7B (on MultiArith) creating and solving its own few-shot examples instead of solving the given problem.}
\begin{tcolorbox}[width=\textwidth, colback=white, colframe=black, boxsep=5pt, arc=2pt]
\textbf{Q:} Paige had 43 math problems and 12 science problems for homework. If she finished 44 of the problems at school, how many problems did she have to do for homework?

\textbf{A:} "1. 43 2. 44 3. 45 4. 46 \\  \#\#\# Watch This \\  Watch this video for help with the concepts in this section. \\  \#\#\# Guidance \\  In the last section, we learned how to solve equations with fractions. In this section, we will learn how to solve equations with decimals. \\  \#\#\#\# Example A \\  Solve the equation for $x$ . \\  $0.2x+0.3=0.5$ \\  Solution: \\  We can solve this equation by subtracting 0.3 from both sides. \\  $0.2x+0.3-0.3=0.5-0.3$ \\  $0.2x=0.2$ \\  Now we can divide both sides by 0.2. \\  $\frac{0.2x}{0.2}=\frac{0.2}{0.2}$ \\  $x=1$ \\  \#\#\#\# Example B \\  Solve the equation for $x$ . \\  $0.05x+0.02=0.07$ \\  Solution: \\  We can solve this equation by subtracting 0.02 from both sides. \\  $0.05x+0.02-0.02=0.07-0.02$ \\  $0.05x=0.05$ \\  Now we can divide both sides by 0.05. \\  $\frac{0.05x}{0.05}=\frac{0.05}{0.05}$ \\  $x=1$ \\  \#\#\#\# Example C \\  Solve the equation for $x$ . \\  $0.005x+0.002=0.007$  \\ Solution: \\  We can solve this equation by subtracting 0.002 from both sides. \\  $0.005x+0.002-0.002=0.007-0.002$ \\  $0.005x=0.005$ \\  Now we can divide both sides by 0.005 \\ 

\end{tcolorbox}
\end{table}

\clearpage
\section{Artifact Licenses}
\label{appendix:artifact_licenses}

All artifacts employed in this study (i.e., models and datasets) are strictly governed by their respective license agreements. Specifically, models such as LLaMA-3.1-8B, Mistral-7B-v0.3, Gemma-7B, and Falcon3-10B-Base are subject to the terms and conditions set forth by their creators, which include usage restrictions, redistribution rights, and potential attribution requirements. For instance, some models may allow non-commercial research use but restrict commercial applications.

Likewise, the datasets utilized (e.g., GSM8K, MultiArith, StrategyQA, BigBench-Disambiguation QA, and BigBench-Logical Deduction) are released under open-access or research-oriented licenses. Researchers are advised to review any additional clauses that detail citation requirements, limitations on commercial use, or derivative works.

We confirm that all experiments in this work were conducted in strict compliance with these licenses. Any redistributions of the models or datasets, as well as any derivative works, must similarly adhere to the terms specified by the original authors and/or dataset providers. Detailed license information for each artifact can be found in the respective original publications and/or in the supplementary materials accompanying this paper.

\end{document}